\pdfoutput=1
\documentclass[sigconf]{acmart}
\usepackage{breqn}
\usepackage{algorithm}
\usepackage{algpseudocode}
\usepackage{comment}
\usepackage[detect-all]{siunitx} 
\usepackage[capitalise]{cleveref}
\usepackage{balance}
\usepackage{lipsum}
\usepackage{bm}
\usepackage{multirow}
\usepackage{makecell}
\usepackage{nth}
\usepackage{subcaption}

\algnewcommand\algorithmicnotation{\textbf{ Notation:}}
\algnewcommand\Notation{\item[\algorithmicnotation]}

\algnewcommand\algorithmicparam{\textbf{ Parameter:}}
\algnewcommand\Param{\item[\algorithmicparam]}

\algnewcommand\algorithmickwin{\textbf{ Input:}}
\algnewcommand\KwIn{\item[\algorithmickwin]}

\algnewcommand\algorithmickwout{\textbf{ Output:}}
\algnewcommand\KwOut{\item[\algorithmickwout]}

\algdef{SE}{Begin}{End}{\textbf{begin}}{\textbf{end}}

\DeclareSIUnit{\OPS}{OPS}

\usepackage{todonotes}

\newcommand*\mean[1]{\overline{#1}}

\DeclareMathOperator*{\argmin}{arg\,min}

\AtBeginDocument{%
  }

\setcopyright{acmcopyright}
\copyrightyear{2023}
\acmYear{2023}
\acmDOI{XXXXXXX.XXXXXXX}

\acmConference[ASP-DAC 2023]{ACM Asia South Pacific Design Automation Conference}{Janurary 16--19,
  2023}{Tokyo, Japan}
\acmPrice{15.00}
\acmISBN{978-1-4503-XXXX-X/18/06}

\begin{document}


\title{Block-Wise Dynamic-Precision Neural Network Training Acceleration via Online Quantization Sensitivity Analytics}

\settopmatter{authorsperrow=4}

\author{Ruoyang Liu}
\authornote{Contributed equally to this research.}
\orcid{0000-0001-9873-6574}
\author{Chenhan Wei}
\authornotemark[1]
\orcid{0000-0003-2683-9835}
\author{Yixiong Yang}
\authornotemark[1]
\orcid{0000-0002-3035-709X}

\affiliation{%
  \institution{Tsinghua University}
  \city{Beijing}
  \country{China}
  \postcode{100084}
}

\author{Wenxun Wang}
\orcid{0000-0002-9213-5257}
\affiliation{%
  \institution{Tsinghua University}
  \city{Beijing}
  \country{China}
  \postcode{100084}
}

\author{Huazhong Yang}
\orcid{0000-0003-2421-353X}
\affiliation{%
  \institution{Tsinghua University}
  \city{Beijing}
  \country{China}
  \postcode{100084}
}

\author{Yongpan Liu}
\email{ypliu@tsinghua.edu.cn}
\orcid{0000-0002-4892-2309}
\authornote{Corresponding author.}
\affiliation{%
  \institution{Tsinghua University}
  \city{Beijing}
  \country{China}
  \postcode{100084}
}

\renewcommand{\shortauthors}{Trovato et al.}

\begin{abstract}
  Data quantization is an effective method to accelerate neural network training and reduce power consumption. However, it is challenging to perform low-bit quantized training: the conventional equal-precision quantization will lead to either high accuracy loss or limited bit-width reduction, while existing mixed-precision methods offer high compression potential but failed to perform accurate and efficient bit-width assignment. In this work, we propose DYNASTY, a block-wise dynamic-precision neural network training framework. DYNASTY provides accurate data sensitivity information through fast online analytics, and maintains stable training convergence with an adaptive bit-width map generator. Network training experiments on CIFAR-100 and ImageNet dataset are carried out, and compared to 8-bit quantization baseline, DYNASTY brings up to $5.1\times$ speedup and $4.7\times$ energy consumption reduction with no accuracy drop and negligible hardware overhead.
\end{abstract}



\maketitle

\section{Introduction}
Neural networks(NN) have been widely adopted in many fields, including computer vision\cite{NIPS2012_c399862d}, speech recognition\cite{6857341}, and natural language processing\cite{NIPS2017_3f5ee243}. However, training NN usually needs large computing power, which requires high-cost cloud servers and makes it infeasible to train on low-power edge devices. Researchers have been studying various methods for NN training acceleration, such as exploiting data pruning algorithm\cite{9218710} and developing more efficient hardware accelerators\cite{9499944}.

As NN training is usually carried out with 32-bit float point data, low-bit network quantization can be an effective way for NN training acceleration. There are 3 ways for NN quantization: post-training quantization\cite{NEURIPS2019_c0a62e13}, quantization aware training\cite{Jacob_2018_CVPR}, and fully-quantized training\cite{2016arXiv160606160Z}. The first 2 methods only quantize network inference data and do not accelerate training process, and they usually need to train auxiliary NN or use methods like evolutionary search, leading to even longer training time\cite{Wang_2020_CVPR}. Meanwhile, fully-quantized training aims at online quantization of all training data, which can significantly reduce the computation cost needed for training.

\begin{figure}[t]
    \centering
    \includegraphics[width=\linewidth]{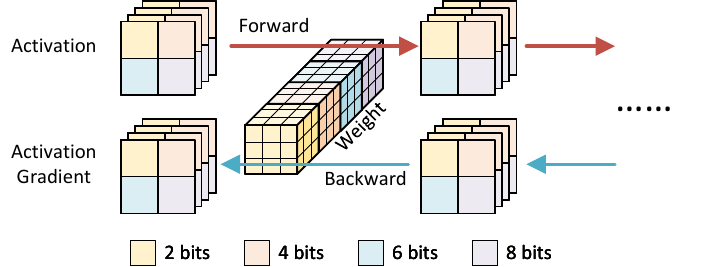}
    \caption{Mixed-precision quantization in NN training}
    \label{fig:mixed-precision}
\end{figure}

Early works on fully-quantized training adopt an equal bit-width for the entire network\cite{2016arXiv160606160Z}, which leads to either high accuracy loss or limited bit-width reduction.
Later works utilize mixed-precision training to provide better acceleration while maintaining network accuracy. Some works like \cite{Dong_2019_ICCV} assign different bit-widths to each network layer, and state-of-the-art quantization results are acquired through block-wise mixed-precision training\cite{pmlr-v139-chen21z}. As shown in \cref{fig:mixed-precision}, block-wise mixed-precision training divides training data into small blocks, and quantizes them into different bit-widths. 
Whether layer-wise or block-wise, mixed-precision training is challenging in how to perform online precision assignment for each layer/block of data. Some of the existing works use data quantization sensitivity acquiring algorithms such as calculating Hessian eigenvalue and trace\cite{Dong_2019_ICCV,NEURIPS2020_d77c7035}, which are too complicated for hardware implementation; other works utilize heuristic methods like greedy search\cite{pmlr-v139-chen21z}, which does not generate accurate bit-width distribution.

To address the aforementioned weakness, we propose DYNASTY (\textbf{DYN}amic \textbf{A}nalytics of \textbf{S}ensitivi\textbf{TY}), a block-wise dynamic-precision NN training framework. By using analytics methods, DYNASTY performs fast dynamic data bit-width assignment, and achieves average 2-bit quantization with minimum accuracy drop.
Key contributions of the work are concluded as follows:

\begin{itemize}
    \item A software-hardware co-designed block-wise fully-quantized mixed-precision NN training framework named DYNASTY. It provides a block-float-point quantized training algorithm with 0-8~bits dynamic block precision, and extends an online bit-width assignment module to the NN acceleration hardware architecture.
    \item A Relative Quantization Sensitivity Analytics algorithm. By applying relaxed Lagrange duality, it transforms the NP-hard bit-width searching problem into $O(N)$ complexity, providing fast but still accurate online data quantization sensitivity analysis.
    \item An Adaptive Bit-Width Map Generator that maps data sensitivity to bit-width. It performs online tuning of algorithm hyperparameter to maintain stable average bit-width, and employs bit-width map temporal smoothing to furtherly enhance network accuracy. 
    \item Experiments on our algorithm show $5.1\times$ training speedup with no accuracy drop on CIFAR-100, and $1.9\times$ speedup with \qty{0,39}{\percent} accuracy drop on ImageNet, compared to 8-bit quantization baseline. Our hardware architecture shows minimum overhead on die area (\qty[retain-explicit-plus]{+1,4}{\percent}) and brings at most \qty{78,8}{\percent} energy consumption reduction.
\end{itemize}

\section{Preliminary \& Related Works}
\subsection{Quantized NN Training Basics}
\label{sec:nn-training}

While NN inference requires only one forward pass computation, training involves three different stages: forward, backward and weight update. Each stage includes multiply-accumulate(MAC) operations on different data: for the forward stage, they are activations and weights; for the backward stage, they are activation gradients (also called errors) and weights; and for the backward stage, they are activations and activation gradients.

Researchers have found training quantization to be more challenging than inference. Xiao Sun et al.\ \cite{NEURIPS2019_65fc9fb4} point out that quantization saturation can be severe in training due to the large dynamic range of gradient data in the backward stage. Later work\cite{NEURIPS2020_13b91943} furtherly discover that different
network layers show widely different ranges of gradients across training epochs.

To better suit these different data distributions, either handcrafted quantization formats for different layers and stages need to be carefully designed as having been done in \cite{NEURIPS2019_65fc9fb4,NEURIPS2020_13b91943}, or better we employ mixed-precision quantization that adaptively adjusts the quantization method for different parts of data in networks.

\subsection{Mixed-Precision NN Training}

Early works of mixed-precision fully-quantized training such as HAWQ\cite{Dong_2019_ICCV} and AdaQS\cite{9054164} only calculate a relative ordering of the importance of each data block. They require empirically assigning bit-width to each block, making their method design include trial and error and hard to be accurate.

A better way to get bit-width distribution is through analytics methods, where the bit-width distribution solving is formulated as an optimization problem to minimize impacts brought by quantization noise. ActNN\cite{pmlr-v139-chen21z} proposes to define the mean square of weight gradients as the optimization target, but this optimization is hard to solve and they have to use greedy search for a heuristic result. HAWQ-V3\cite{pmlr-v139-yao21a} needs to solve an integer linear programming problem, and MPQCO\cite{Chen_2021_ICCV} formulates it into a multiple-choice knapsack problem, both of which are NP-hard to solve.

\section{Block-Wise Dynamic-Precision NN Training Framework}
\subsection{Quantization Training Framework}
\begin{figure}[t]
    \centering
    \includegraphics[width=0.8\linewidth]{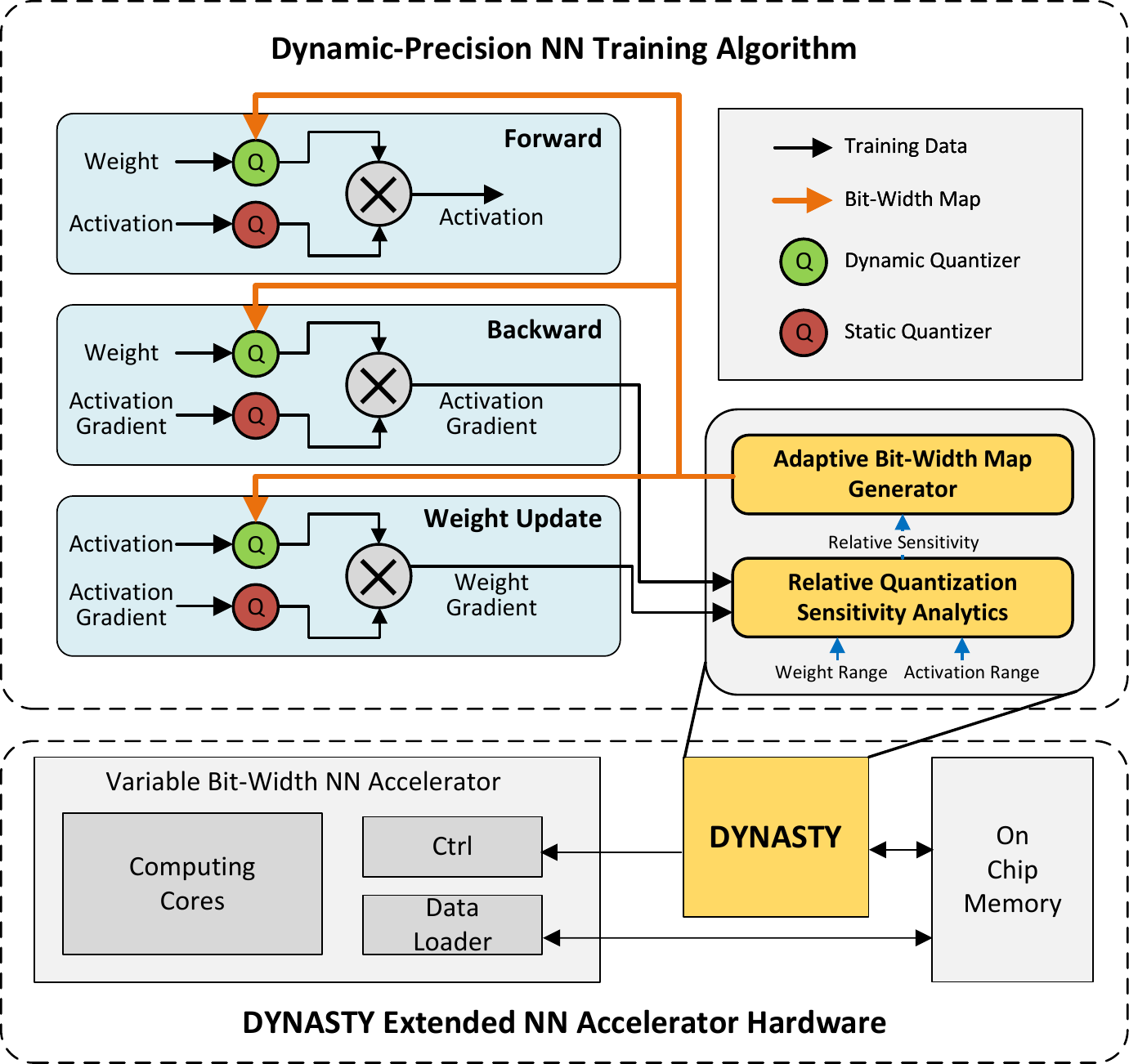}
    \caption{Proposed block-wise dynamic-precision NN training framework}
    \label{fig:framework}
\end{figure}
\Cref{fig:framework} depicts the overall framework of our system, where the DYNASTY module extends an NN accelerator for block-wise dynamic-precision training. Input data of all training stages are quantized by Block-Float-Point(BFP) format, which divides data into $4\times4$ blocks\footnote{Block division is along $C_{in}$ and $C_{out}$ dimensions for weight and weight gradients, $C_{in}$ and $Batch$ dimensions for activations, and $C_{out}$ and $Batch$ dimensions for activation gradients.}. Data in each block have mantissa quantized into a same bit-width, and they also share a single 8-bit exponent. 
Two strategies are used for mantissa bit-width assignment:
in each training stage, one of the input data is statically quantized to 8-bit mantissa; sensitivity analytics is performed for the other input data, whose mantissa is dynamically quantized into 0-8 bits according to the bit-width map generated by the DYNASTY module.

The DYNASTY module is in charge of the dynamic bit-width assignment, and consists of two sub-modules. The first sub-module, named \textbf{Relative Quantization Sensitivity Analytics}, receives the current quantization range together with data gradients, and generates a relative quantization sensitivity value for each block. This analytics happens at the end of the training of each mini-batch. This sub-module will be detailed in \cref{sec:relative_quantization_sensitivity_analytics}.
The second sub-module, the \textbf{Adaptive Bit-Width Map Generator}, works at the end of each training epoch. It maps the sensitivities into a bit-width distribution and maintains stable training. This mapping happens at the end of each training epoch. Details of this sub-module will be provided in \cref{sec:high_efficiency_adaptive_bit_width_map_updator}.

\subsection{Relative Quantization Sensitivity Analytics}
\label{sec:relative_quantization_sensitivity_analytics}
To perform fast online data analysis and dynamically identify which data is more important and need larger quantization bit-width than others, the Relative Quantization Sensitivity Analytics module is designed to calculate data quantization sensitivity based on their quantization range and corresponding gradients. We first establish an optimization problem based on the minimization of first-order mean square loss noise. Then we propose to solve the optimization by relaxed Lagrange duality method and get the sensitivity values.

\begin{figure}[t]
    \centering
    \includegraphics[width=\linewidth]{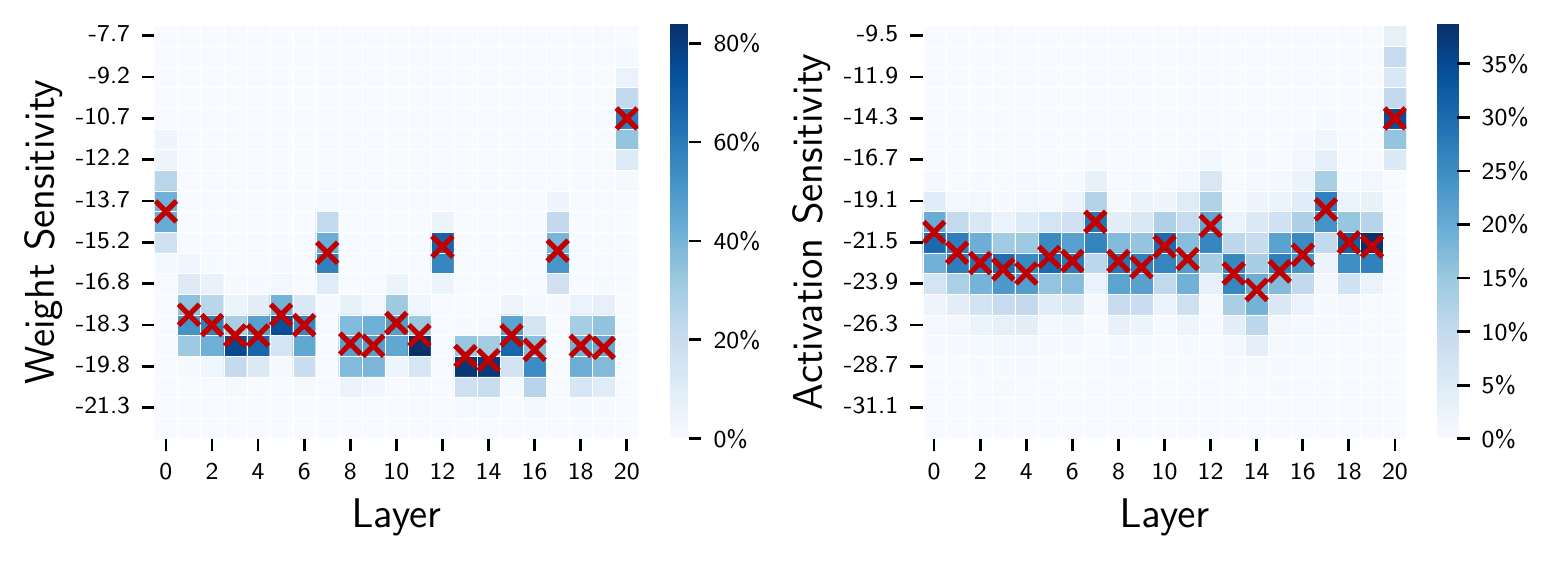}
    \caption{An example of per-layer relative sensitivity distribution. Red markers represent the mean sensitivity per layer, and different colored squares represent the proportion of different sensitivity ranges in each layer.}
    \label{fig:sensitivity-dist}
\end{figure}

\subsubsection{Optimization Problem Based On First-Order Mean Square Loss Noise}
Quantization brings noise to network training process, leading to less stable training and lower network accuracy. An ideal bit-width assignment should minimize this noise and bring minimum perturbation to the train loss. We propose to use the first-order mean square loss noise($\Delta L^2$) to quantify this perturbation. The math formula of $\Delta L^2$ caused by weight quantization is shown as \cref{eq:noise_weight}. It is calculated as the mean square value of first-order loss noise caused by each quantized data. And we use $S_i$ to denote the ratio of loss noise to quantization levels in each data block. In \cref{eq:noise_weight}, $\Delta W_{i,j}$ is the gradient of the $j$th weight of the $i$th block; $\sigma_{i,j}$ is the quantization noise of corresponding data; $N$ is the total number of data blocks; $B$ is the number of values inside each block; $b_i$ is the quantization bit-width of the mantissa of the $i$th block of weights; and $e_{i}$ is the exponent of each block so that $2^{e_i}$ is the corresponding quantization range. Note that, $\Delta L^2$ brought by activation quantization can also be calculated in the same way, and we omit its formula here for simplicity. With our formulation, $\Delta L^2$ is calculated from the quantization range of each block and their corresponding gradient values, both of which are already known during NN training.

\begin{equation}
    \label{eq:noise_weight}
    \begin{alignedat}{2}
        &\Delta L^2 && = \frac{1}{N B}\sum_{i=1}^{N} \sum_{j=1}^{B} \left( \Delta W_{i,j} \cdot \sigma_{i,j} \right)^2 = \frac{2}{3N B} \sum_{i=1}^{N}S_{i} \left(2^{-b_{i}}\right)^2 \\
        & S_{i}     && = \frac{\left(2^{e_{i}}\right)^2}{16}\sum_{j=1}^{B} \Delta W_{i,j}^2 \\
    \end{alignedat}
\end{equation}

The full optimization is shown as \cref{eq:opt}. Aside from $\Delta L^2$ as our optimization target function, we also specify the optimization constraints. The $\alpha$ denotes the desired average computation bit-width, which is the $T_i$ weighted average of block bit-width $b_i$, where $T_i$ represents the amount of computation associated with corresponding data. Bit-width lower bound $0$ and upper bound $\beta$ are also specified.

\begin{subequations}
    \label{eq:opt}
    \begin{alignat}{2}
        &\!\argmin_{\bm{b}}       & \qquad& f\!\left(\bm{b}\right) = \Delta L^2 \\
        &\text{subject to} &      & \frac{\sum_{i=1}^{N} {T_{i} b_{i}}}{\sum T_{i}}  \leq \alpha \label{eq:calc_num} \\
        &                  &      & b_{i} \in \mathbb{Z} \label{eq:integer_set} \\
        &                  &      & 0 \leq b_{i} \leq \beta \label{eq:min_max_bits}
    \end{alignat}
\end{subequations}

\subsubsection{Problem Solving for Relative Quantization Sensitivity}

Optimization in \cref{eq:opt} is a mixed integer convex programming problem. It is NP-hard to solve and too complicated for hardware implementation. Thus we propose to relax the restriction in \cref{eq:integer_set} to the $\mathbb{R}$ set, and then apply Lagrange duality to this problem. This leads to the solution shown in \cref{eq:opt_result}. The bit-width is calculated from value $\bm{r}$ and a hyperparameter $\lambda$. $\bm{r}$ is easy to compute based on \cref{eq:r}. It is positively correlated with $S_i$ and negatively correlated with $T_i$, and only requires $O(N)$ time complexity for weights and activations of the entire network. $\bm{r}$ tells the relative bit-width between data blocks, and we call it the \textbf{relative quantization sensitivity}. Though we still do not know the value of $\lambda$, \textbf{we successfully transform the searching of bit-width of N different blocks into the searching of only a single hyperparameter.} We call $\lambda$ the \textbf{global quantization
coefficient}.

\begin{subequations}
    \label{eq:opt_result}
    \begin{alignat}{2}
        & b_{i} && = \max\!\left(0,\min\!\left(r_{i} - \lambda, \beta\right)\right) \label{eq:b}\\
        & r_{i} && = \frac{1}{2}\log_2\frac{S_{i}}{T_{i}} \label{eq:r}
    \end{alignat}
\end{subequations}

An example of $\bm{r}$ distribution during training of ResNet-18 is shown as \cref{fig:sensitivity-dist}. We can see that the desired bit-width distribution should be different across network layers. Data in several layers requires much larger bit-widths than others, including weights in the first, the last, and the 3 shortcut layers, and activation in the last layer. Bit-width distribution inside each layer is also different, and activations' bit-widths tend to be more spread out and reside in larger ranges than weights'. These observations suggest that manually setting bit-width can hardly suit the data requirement, and online bit-width assignment based on data sensitivity is needed. 

\begin{figure}[t]
    \centering
    \includegraphics[width=\linewidth]{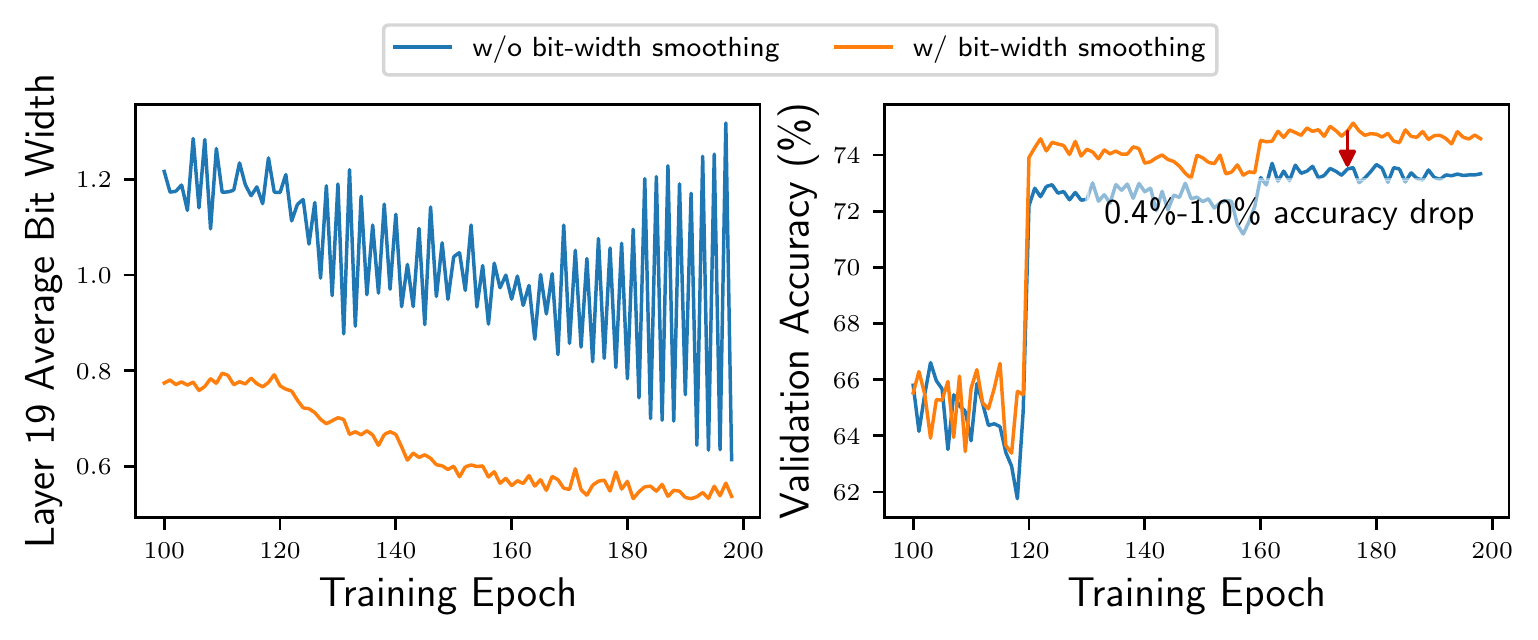}
    \caption{Data bit-width fluctuation reduces network accuracy.}
    \label{fig:bw_smooth}
\end{figure}

However, though \cref{eq:r} uncovers the relative bit-widths between data blocks, it still does not give us the absolute bit-width of each block of data. Two problems remain to be solved. The first is how to search the value of $\lambda$. The second is that the sensitivities of some blocks are unstable across training epochs, leading to bit-width fluctuating periodically as shown in \cref{fig:bw_smooth}, which brings \qty{0,4}{\percent} to \qty{1}{\percent} network accuracy drop in our experiment. Both of these problems are addressed in \cref{sec:high_efficiency_adaptive_bit_width_map_updator}.

\subsection{Adaptive Bit-Width Map Generator}
\label{sec:high_efficiency_adaptive_bit_width_map_updator}
The Adaptive Bit-Width Map Generator aims to solve the remaining problems mentioned in \cref{sec:relative_quantization_sensitivity_analytics} and maps the relative sensitivities $\bm{r}$ of each block to their exact quantization bit-widths.

\subsubsection{Dertermine the Value of $\lambda$}
To find the exact global quantization coefficients $\lambda$, we develop an adaptive quantization coefficient adjustment algorithm to dynamically determine the coefficient during training.
In theory, if we ignore these upper and lower bounds constraints in \cref{eq:b}, an approximate of $\lambda$ can be easily deduced as $\hat{\lambda} = \sum_i B_i - N\alpha$, where $N$ is the number of blocks. With experiments, we discover that in the initial training of a network, differences in the importance of data in the network are not much and seldomly make $b_i$ reach the lower or upper bound, thus $\hat{\lambda}$ is a good estimation of the initial $\lambda$ value.

\begin{figure}[t]
    \centering
    \includegraphics[width=\linewidth]{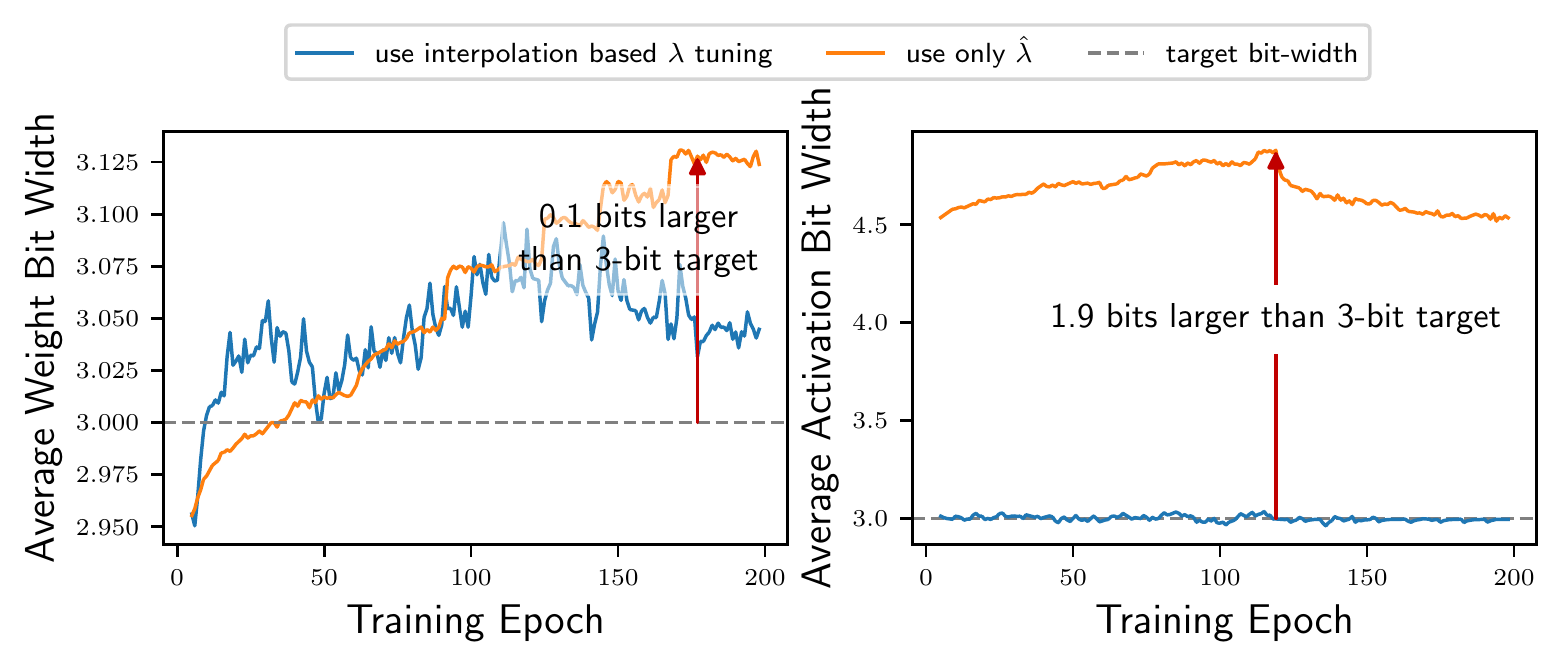}
    \caption{Real network bit-width when targeting 3-bit quantization. Without linear interpolation-based tuning of $\lambda$, the average bit-width tends to become larger than desired.}
    \label{fig:k_tuning}
\end{figure}

However, as shown in \cref{fig:k_tuning}, using only $\hat{\lambda}$ leads to the average bit-width becoming larger than desired as training continues, especially for activation data. To prevent $\lambda$ drifting away from the desired value, we design \cref{alg:k_tuning}, which tunes $\lambda$ with linear interpolation using the last $\lambda$ value as the start point of the interpolation. The number of interpolation iterations $L$ is empirically set to 3. By using this method, average bit-width can be maintained stably at the target bit-width, as the blue lines show in \cref{fig:k_tuning}.

\begin{algorithm}[tb]
    \caption{Global Quantization Coefficient Tuning}
    \label{alg:k_tuning}   
    \begin{algorithmic}[1]
        \KwIn $\lambda_{in}$: the last global quantization coefficient
        \Statex $\bm{r}$: relative sensitivity
        \Statex $\alpha$: target averaged quantization bit-width
        \KwOut $\lambda_{out}$: the tuned global quantization coefficient
        \Param $L$: number of interpolation iterations
    \Begin
        \State $\lambda_{out} = \lambda_{in}$
        \State $\lambda_0, \lambda_1 \gets \min\!\left(\bm{r}\right) - \alpha, \max\!\left(\bm{r}\right) - \alpha$
        \For{$l = 1$ to $L$}
        \State $b_0 \gets \frac{1}{N} \sum_i \max\!\left(\min\!\left(r_i - \lambda_0, \beta\right), 0\right)$
        \State $b_1 \gets \frac{1}{N} \sum_i \max\!\left(\min\!\left(r_i - \lambda_1, \beta\right), 0\right)$
        \State $\mean{b} \gets \frac{1}{N} \sum_i \max\!\left(\min\!\left(r_i - \lambda, \beta\right), 0\right)$
            \If{$\mean{b} > \alpha$}
                \State $t \gets \left(\mean{b} - \alpha\right) / \left(\mean{b} - b_1\right)$
                \State $\lambda_0, \lambda_{out} \gets \lambda_{out}, \left(1 - t\right)\lambda_{out} + t \cdot \lambda_1$
            \Else
                \State $t \gets \left(\mean{b} - \alpha\right) / \left(\mean{b} - b_0\right)$
                \State $\lambda_{out}, \lambda_1 \gets \left(1 - t\right)\lambda_{out} + t \cdot \lambda_0, \lambda_{out}$
            \EndIf
        \EndFor
    \End
    \end{algorithmic}
\end{algorithm}

\subsubsection{Prevent Bit-Width Fluctuatation}
To solve the bit-width fluctuation problem, we propose a bit-width map temporal smoothing method: at every epoch, an exponential moving average is performed on the bit-width map, shown as \cref{eq:smooth}. This prevents sudden changes in data bit-width caused by any calculation error. The smooth factor $\gamma$ is empirically set to \num{0,5}.
\begin{dmath}
    \label{eq:smooth}
    b_{i,\text{smoothed}} = \gamma \cdot b_{i,\text{last}} + \left( 1 - \gamma \right) b_{i,\text{new}}
\end{dmath}

Note that, during the first training epoch when we have not got the value of $\hat{\lambda}$, we assign 4~bits to every block as the initial bit-width map. And for hardware simplicity, we perform rounding on the bit-width map to only use 0,2,4,6,8 as possible quantized bit-widths. Specially, when a block is quantized to 0~bit, we skip the computation of this block. The smoothed bit-width map is then sent to the accelerator computing cores for dynamic data quantization.

\section{Hardware Architecture}

\begin{figure}[t]
    \centering
    \includegraphics[width=\linewidth]{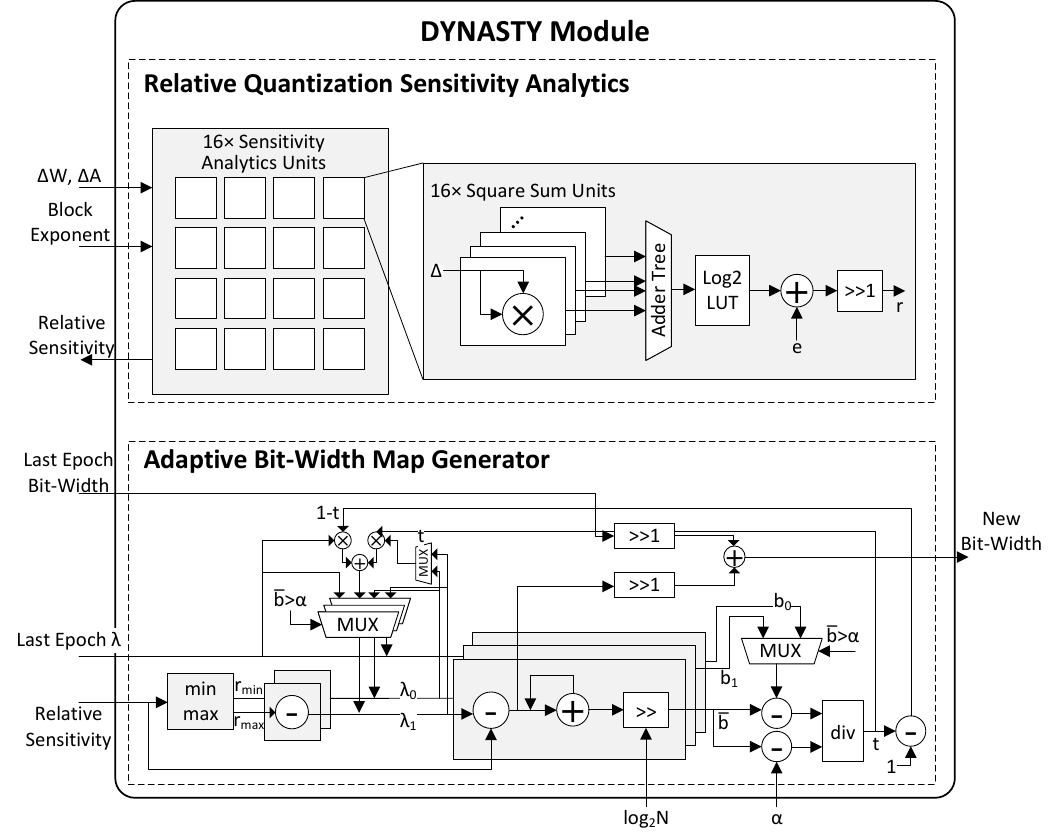}
    \caption{DYNASTY module hardware architecture}
    \label{fig:hardware_inside}
\end{figure}

Support of variable bit-width operation has been realised through NN accelerators with bit-serial-based computation array\cite{8481682,9499944}. These accelerators get linear speedup as the operation bit-width shrinks.
The DYNASTY hardware architecture extends such accelerators with online bit-width assignment ability.

\cref{fig:hardware_inside} shows the hardware architecture of the DYNASTY module. The use of 16 parallel sensitivity analytics units makes it capable of calculating the sensitivities and generating bit-width map for 16 blocks, i.e.\ 256 weights or activations in one cycle. Note that, the DYNASTY module is running in parallel with the main array of the accelerator. A recovery remainder-based divider is used in the Bit-Width Map Generator, but since there is only one divider, and the divisions only happen at the end of each training epoch, it does not bring much area and power overhead.

\section{Experimental Results}
\subsection{Experiment Setup}
In our experiments, we train ResNet-18 network\cite{He_2016_CVPR} on the CIFAR-100\cite{cifar} and ImageNet\cite{5206848} dataset. For both datasets, we adopt a batch size of 128, and use SGD as the optimizer with an initial learning rate of \num{0,1}. For CIFAR-100, we train the network for 200 epochs; learning rate linear warming up is used in the \nth{1} epoch; and we perform learning rate decay at the \nth{60}, \nth{120}, and \nth{160} epochs with a decay factor of \num{0,2}. For ImageNet, we train the network for 95 epochs; learning rate linear warming up is used in the first 5 epochs; and we perform learning rate decay at the \nth{30}, \nth{60}, and \nth{80} epoch with a decay factor of \num{0,1}.

We also perform training with 2 other methods aside from DYNASTY as our baseline. The first method is to train with float-32 format without quantization; the second is 8-bit quantized training according to the work of Ron Banner et al.\ \cite{NEURIPS2018_e82c4b19}.

We implement the DYNASTY module in Verilog and get its area and power information by synthesizing it in Synopsys
Design Compiler with TSMC 65nm CMOS technology. The hardware metrics of other modules in a full accelerator are taken from the UNPU accelerator\cite{8481682}.

\begin{table}[t]
    \caption{Network accuracy with different average quantization bit-width}
    \label{tab:accuracy}
    \begin{tabular}{cccc}
        \toprule
        Dataset & Method & \thead{Average \\ Bit-Width$^{*}$} & \thead{Top 1 Validation \\ Accuracy (\%)} \\
        \midrule
        \multirowcell{7}{CIFAR-100} 
            & Float-Point & All Float-32      & \num{76,54}              \\
        \cmidrule(l){2-4}
            & Banner et al. & 8 & \num{74,77} (\num{-1,77}) \\
        \cmidrule(l){2-4}
            & \multirowcell{4}{DYNASTY}
                          & 6  & \num{76,57} (\num[retain-explicit-plus]{+0,03})  \\
            &             & 4  & \num{76,55} (\num[retain-explicit-plus]{+0,01}) \\
            &             & 3  & \num{75,81} (\num{-0,73}) \\
            &             & 2  & \num{75,14} (\num{-1,40}) \\
        \midrule
        \multirowcell{7}{ImageNet} 
            & Float-Point & All Float-32      & \num{68,77}              \\
        \cmidrule(l){2-4}
            & Banner et al. & 8 & \num{68,28} (\num{-0,49}) \\
        \cmidrule(l){2-4}
            & \multirowcell{4}{DYNASTY}
                          & 6  & \num{68,60} (\num{-0,17})  \\
            &             & 4  & \num{67,89} (\num{-0,88}) \\
            &             & 3  & \num{65,84} (\num{-2,93}) \\
            &             & 2  & \num{62,52} (\num{-6,25}) \\
    \bottomrule
    \multicolumn{4}{l}{\makecell[l]{$^{*}$Average bit-width refers to the bit-width of dynamic \\ quantized training data.}}
  \end{tabular}
\end{table}

\subsection{Algorithm Evaluation}
\subsubsection{Network Accuracy}

\Cref{tab:accuracy} shows the validation accuracy that DYNASTY gets with different quantization bit-width. 
On CIFAR-100, we achieve 4-bit quantization with no accuracy drop compared to float-32 training; even with 2-bit quantization, we only get an accuracy drop of \qty{1,40}{\percent} compared to float-32 training and get better accuracy than the 8-bit baseline. On ImageNet, while the accuracy drop with 2-bit quantization is not small (\qty{6,25}{\percent}), the 4-bit training experiment still reveals good accuracy with only \qty{0,88}{\percent} accuracy drop.

\subsubsection{Training Convergence Speed Improvement}
\begin{figure}[t]
    \centering
    \includegraphics[width=\linewidth]{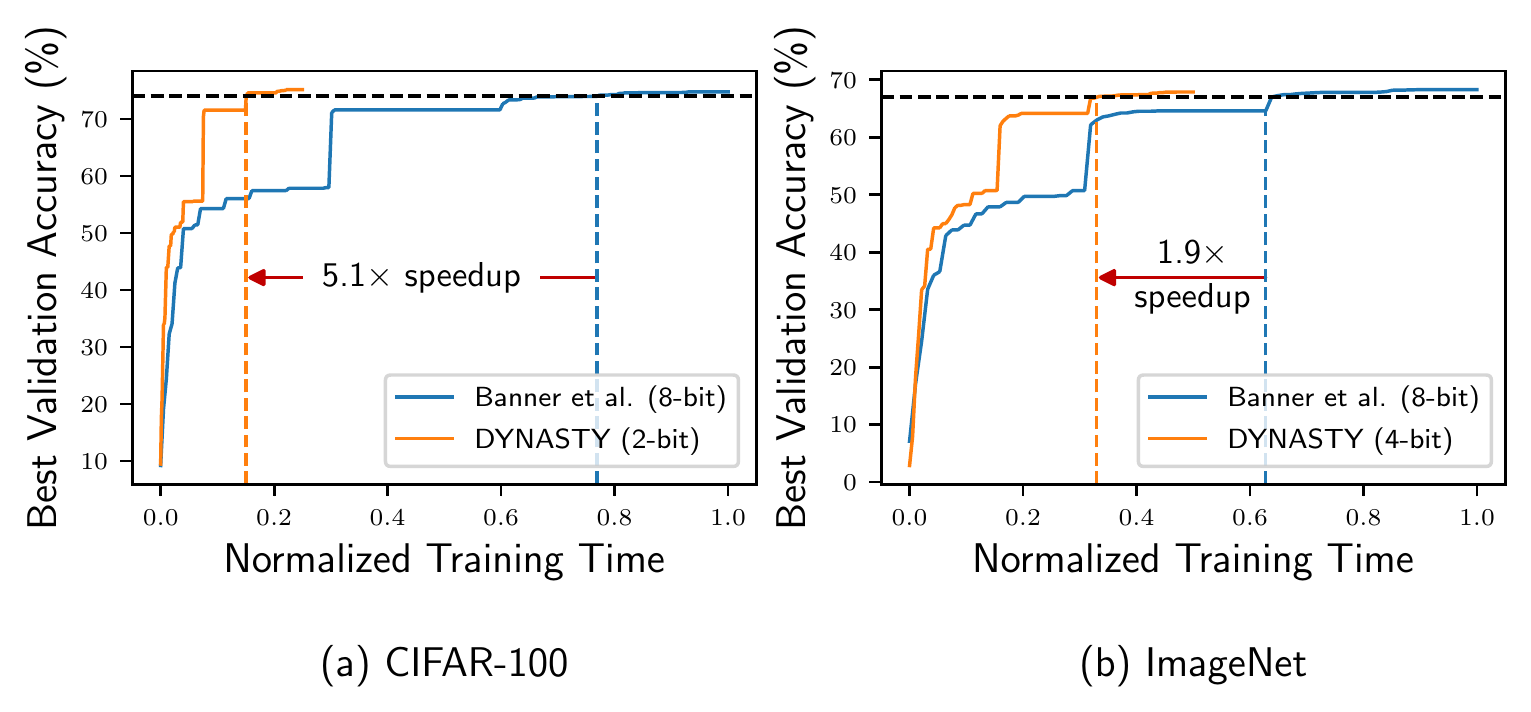}
    \caption{Training convergence speed improvement with DYNASTY}
    \label{fig:convergence}
\end{figure}

\Cref{fig:convergence} presents the accuracy-time curve in the experiments. Note that, the training speed for unquantized float-32 training is not compared here, as the cores of the UNPU accelerator cannot handle float-32 operations.

Because the DYNASTY module runs in parallel with the accelerator cores, we enjoy a linear speedup as the quantization bit-width decreases. 
Compared to 8-bit training baseline, for CIFAR-100, DYNASTY reaches the same validation accuracy (\qty{74}{\percent}) in \qty{58,3}{\percent}, \qty{38,9}{\percent}, \qty{29,2}{\percent}, and \qty{19,6}{\percent} training time with 6, 4, 3, 2-bit quantization, respectively; because the better convergence than baseline, the speedup ($5.1\times$ with 2-bit) is even higher than expectation ($4\times$).
For ImageNet, DYNASTY needs \qty{75,0}{\percent} and \qty{52,5}{\percent} time to get \qty{67}{\percent} accuracy with 6 and 4-bit quantization, respectively.

\subsection{Architecture Evaluation}
\begin{table}
    \caption{Area and power overhead of DYNASTY}
    \label{tab:overhead}
    \begin{tabular}{ccc}
      \toprule
       Module & \thead{Area\\(\unit{\milli\metre\squared})} & \thead{Power Consumption\\(\unit{\milli\watt}@\qty{200}{\mega\hertz})} \\
      \midrule
      DYNASTY  & \num{0,219} & \makecell{\num{10,3} (4-bit, ImageNet)} \\
      \makecell{Main Accelerator} & 16 & 297 \\
    \bottomrule
  \end{tabular}
\end{table}

\Cref{tab:overhead} shows the area and power of the DYNASTY module compared to other modules in the accelerator. We can see that DYNASTY brings no more than \qty{1,4}{\percent} area overhead. 

\begin{figure}[t]
    \centering
    \includegraphics[width=\linewidth]{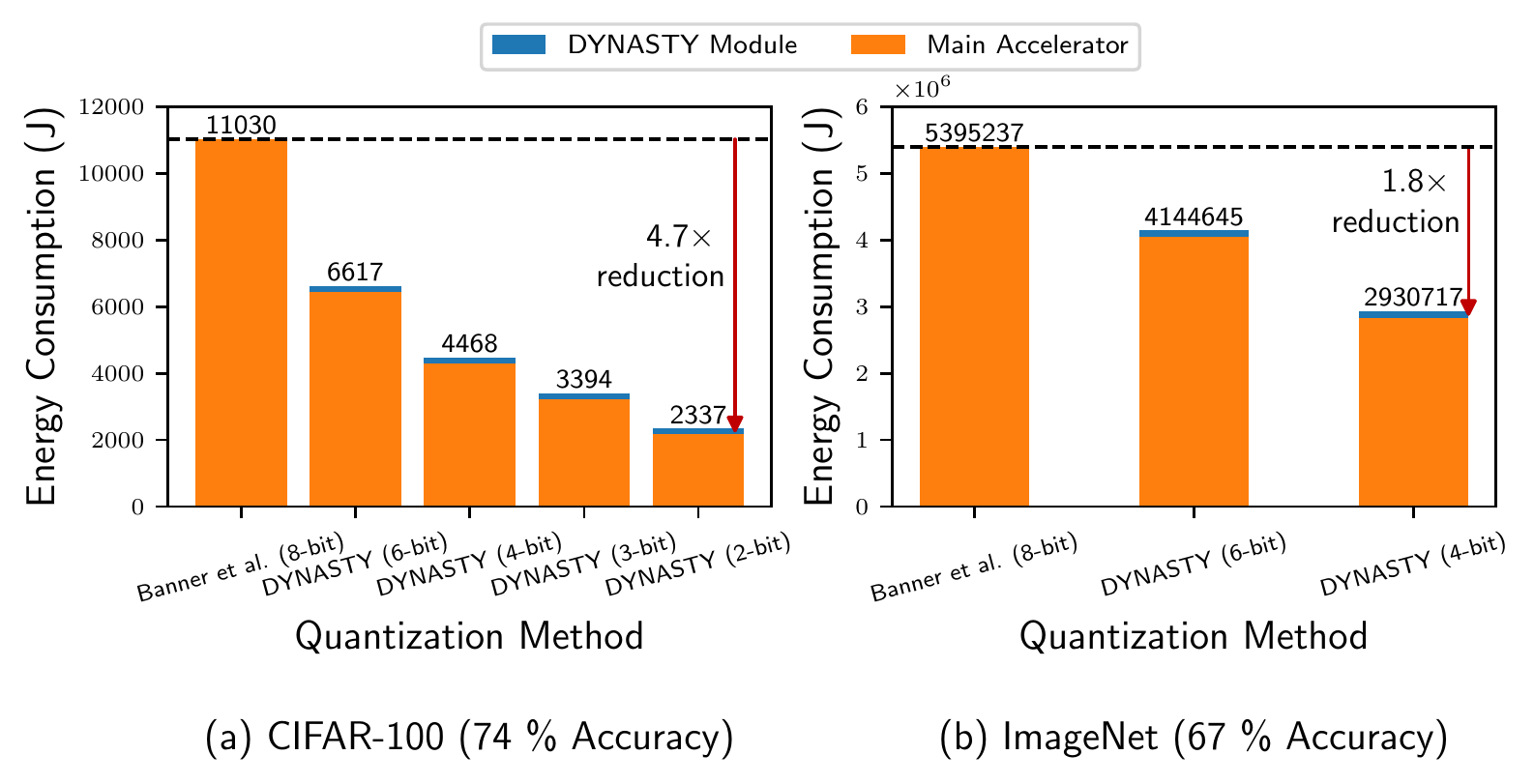}
    \caption{Energy consumption reduction with DYNASTY}
    \label{fig:power}
\end{figure}

The average power consumption of DYNASTY varies with the network scale and quantization bit-width as these factors affect the duty cycle of the DYNASTY module, and we only show the value in a typical situation in \cref{tab:overhead}. Details on the energy consumption overhead can be seen in \cref{fig:power}. DYNASTY brings at most \qty{7,9}{\percent} power consumption overhead to the accelerator, but this overhead is overcome by the energy saving due to shorter training time with lower-bit quantization. When targeting \qty{74}{\percent} accuracy on CIFAR-100, we get \qty{40,0}{\percent}, \qty{59,5}{\percent}, \qty{69,2}{\percent}, and \qty{78,8}{\percent} energy consumption reduction when quantizing to average 6, 4, 3, and 2~bits. For ImageNet, we also achieve \qty{23,2}{\percent} and \qty{45,7}{\percent} energy reduction with 6 and 4-bit quantization when targeting \qty{67}{\percent} accuracy.

\section{Conclusion}
In this paper, we propose DYNASTY, a software-hardware co-designed block-wise dynamic-precision neural network training framework. We utilize an analytics method based on network first-order mean square loss noise to efficiently generate data quantization sensitivity online, and develop an Adaptive Bit-Width Map Generator to accurately map sensitivity to bit-width distribution that maintains desired computation reduction and high network accuracy at the same time. Experiments on CIFAR-100 and ImageNet dataset are carried out, and compared to 8-bit quantization baseline, we achieve up to $5.1\times$ speedup and $4.7\times$ energy consumption reduction with no accuracy drop and negligible hardware overhead.
\balance
\bibliographystyle{unsrtnat}
\bibliography{main.bbl}

\begin{thebibliography}{22}
\providecommand{\natexlab}[1]{#1}
\providecommand{\url}[1]{\texttt{#1}}
\expandafter\ifx\csname urlstyle\endcsname\relax
  \providecommand{\doi}[1]{doi: #1}\else
  \providecommand{\doi}{doi: \begingroup \urlstyle{rm}\Url}\fi

\bibitem[Krizhevsky et~al.(2012)Krizhevsky, Sutskever, and
  Hinton]{NIPS2012_c399862d}
Alex Krizhevsky, Ilya Sutskever, and Geoffrey~E Hinton.
\newblock Imagenet classification with deep convolutional neural networks.
\newblock In F.~Pereira, C.J. Burges, L.~Bottou, and K.Q. Weinberger, editors,
  \emph{Advances in Neural Information Processing Systems}, volume~25. Curran
  Associates, Inc., 2012.
\newblock URL
  \url{https://proceedings.neurips.cc/paper/2012/file/c399862d3b9d6b76c8436e924a68c45b-Paper.pdf}.

\bibitem[Abdel-Hamid et~al.(2014)Abdel-Hamid, Mohamed, Jiang, Deng, Penn, and
  Yu]{6857341}
Ossama Abdel-Hamid, Abdel-rahman Mohamed, Hui Jiang, Li~Deng, Gerald Penn, and
  Dong Yu.
\newblock Convolutional neural networks for speech recognition.
\newblock \emph{IEEE/ACM Transactions on Audio, Speech, and Language
  Processing}, 22\penalty0 (10):\penalty0 1533--1545, Oct 2014.
\newblock ISSN 2329-9304.
\newblock \doi{10.1109/TASLP.2014.2339736}.

\bibitem[Vaswani et~al.(2017)Vaswani, Shazeer, Parmar, Uszkoreit, Jones, Gomez,
  Kaiser, and Polosukhin]{NIPS2017_3f5ee243}
Ashish Vaswani, Noam Shazeer, Niki Parmar, Jakob Uszkoreit, Llion Jones,
  Aidan~N Gomez, \L~ukasz Kaiser, and Illia Polosukhin.
\newblock Attention is all you need.
\newblock In I.~Guyon, U.~Von Luxburg, S.~Bengio, H.~Wallach, R.~Fergus,
  S.~Vishwanathan, and R.~Garnett, editors, \emph{Advances in Neural
  Information Processing Systems}, volume~30. Curran Associates, Inc., 2017.
\newblock URL
  \url{https://proceedings.neurips.cc/paper/2017/file/3f5ee243547dee91fbd053c1c4a845aa-Paper.pdf}.

\bibitem[Dai et~al.(2020)Dai, Yang, Ye, Cheng, Luo, Song, Chen, and
  Zhao]{9218710}
Pengcheng Dai, Jianlei Yang, Xucheng Ye, Xingzhou Cheng, Junyu Luo, Linghao
  Song, Yiran Chen, and Weisheng Zhao.
\newblock Sparsetrain: Exploiting dataflow sparsity for efficient convolutional
  neural networks training.
\newblock In \emph{2020 57th ACM/IEEE Design Automation Conference (DAC)},
  pages 1--6, July 2020.
\newblock \doi{10.1109/DAC18072.2020.9218710}.

\bibitem[Zhao et~al.(2021)Zhao, Liu, Du, Guo, Hu, Zhuang, Zhang, Song, Li,
  Zhang, Li, Xu, and Chen]{9499944}
Yongwei Zhao, Chang Liu, Zidong Du, Qi~Guo, Xing Hu, Yimin Zhuang, Zhenxing
  Zhang, Xinkai Song, Wei Li, Xishan Zhang, Ling Li, Zhiwei Xu, and Tianshi
  Chen.
\newblock Cambricon-q: A hybrid architecture for efficient training.
\newblock In \emph{2021 ACM/IEEE 48th Annual International Symposium on
  Computer Architecture (ISCA)}, pages 706--719, June 2021.
\newblock \doi{10.1109/ISCA52012.2021.00061}.

\bibitem[Banner et~al.(2019)Banner, Nahshan, and Soudry]{NEURIPS2019_c0a62e13}
Ron Banner, Yury Nahshan, and Daniel Soudry.
\newblock Post training 4-bit quantization of convolutional networks for
  rapid-deployment.
\newblock In H.~Wallach, H.~Larochelle, A.~Beygelzimer, F.~d\textquotesingle
  Alch\'{e}-Buc, E.~Fox, and R.~Garnett, editors, \emph{Advances in Neural
  Information Processing Systems}, volume~32. Curran Associates, Inc., 2019.
\newblock URL
  \url{https://proceedings.neurips.cc/paper/2019/file/c0a62e133894cdce435bcb4a5df1db2d-Paper.pdf}.

\bibitem[Jacob et~al.(2018)Jacob, Kligys, Chen, Zhu, Tang, Howard, Adam, and
  Kalenichenko]{Jacob_2018_CVPR}
Benoit Jacob, Skirmantas Kligys, Bo~Chen, Menglong Zhu, Matthew Tang, Andrew
  Howard, Hartwig Adam, and Dmitry Kalenichenko.
\newblock Quantization and training of neural networks for efficient
  integer-arithmetic-only inference.
\newblock In \emph{Proceedings of the IEEE Conference on Computer Vision and
  Pattern Recognition (CVPR)}, June 2018.

\bibitem[{Zhou} et~al.(2016){Zhou}, {Wu}, {Ni}, {Zhou}, {Wen}, and
  {Zou}]{2016arXiv160606160Z}
Shuchang {Zhou}, Yuxin {Wu}, Zekun {Ni}, Xinyu {Zhou}, He~{Wen}, and Yuheng
  {Zou}.
\newblock {DoReFa-Net: Training Low Bitwidth Convolutional Neural Networks with
  Low Bitwidth Gradients}.
\newblock \emph{arXiv e-prints}, art. arXiv:1606.06160, June 2016.

\bibitem[Wang et~al.(2020)Wang, Wang, Cai, Lin, Liu, Wang, Lin, and
  Han]{Wang_2020_CVPR}
Tianzhe Wang, Kuan Wang, Han Cai, Ji~Lin, Zhijian Liu, Hanrui Wang, Yujun Lin,
  and Song Han.
\newblock Apq: Joint search for network architecture, pruning and quantization
  policy.
\newblock In \emph{Proceedings of the IEEE/CVF Conference on Computer Vision
  and Pattern Recognition (CVPR)}, June 2020.

\bibitem[Dong et~al.(2019)Dong, Yao, Gholami, Mahoney, and
  Keutzer]{Dong_2019_ICCV}
Zhen Dong, Zhewei Yao, Amir Gholami, Michael~W. Mahoney, and Kurt Keutzer.
\newblock Hawq: Hessian aware quantization of neural networks with
  mixed-precision.
\newblock In \emph{Proceedings of the IEEE/CVF International Conference on
  Computer Vision (ICCV)}, October 2019.

\bibitem[Chen et~al.(2021{\natexlab{a}})Chen, Zheng, Yao, Wang, Stoica,
  Mahoney, and Gonzalez]{pmlr-v139-chen21z}
Jianfei Chen, Lianmin Zheng, Zhewei Yao, Dequan Wang, Ion Stoica, Michael
  Mahoney, and Joseph Gonzalez.
\newblock Actnn: Reducing training memory footprint via 2-bit activation
  compressed training.
\newblock In Marina Meila and Tong Zhang, editors, \emph{Proceedings of the
  38th International Conference on Machine Learning}, volume 139 of
  \emph{Proceedings of Machine Learning Research}, pages 1803--1813. PMLR,
  18--24 Jul 2021{\natexlab{a}}.
\newblock URL \url{https://proceedings.mlr.press/v139/chen21z.html}.

\bibitem[Dong et~al.(2020)Dong, Yao, Arfeen, Gholami, Mahoney, and
  Keutzer]{NEURIPS2020_d77c7035}
Zhen Dong, Zhewei Yao, Daiyaan Arfeen, Amir Gholami, Michael~W Mahoney, and
  Kurt Keutzer.
\newblock Hawq-v2: Hessian aware trace-weighted quantization of neural
  networks.
\newblock In H.~Larochelle, M.~Ranzato, R.~Hadsell, M.F. Balcan, and H.~Lin,
  editors, \emph{Advances in Neural Information Processing Systems}, volume~33,
  pages 18518--18529. Curran Associates, Inc., 2020.
\newblock URL
  \url{https://proceedings.neurips.cc/paper/2020/file/d77c703536718b95308130ff2e5cf9ee-Paper.pdf}.

\bibitem[Sun et~al.(2019)Sun, Choi, Chen, Wang, Venkataramani, Srinivasan, Cui,
  Zhang, and Gopalakrishnan]{NEURIPS2019_65fc9fb4}
Xiao Sun, Jungwook Choi, Chia-Yu Chen, Naigang Wang, Swagath Venkataramani,
  Vijayalakshmi~(Viji) Srinivasan, Xiaodong Cui, Wei Zhang, and Kailash
  Gopalakrishnan.
\newblock Hybrid 8-bit floating point (hfp8) training and inference for deep
  neural networks.
\newblock In H.~Wallach, H.~Larochelle, A.~Beygelzimer, F.~d\textquotesingle
  Alch\'{e}-Buc, E.~Fox, and R.~Garnett, editors, \emph{Advances in Neural
  Information Processing Systems}, volume~32. Curran Associates, Inc., 2019.
\newblock URL
  \url{https://proceedings.neurips.cc/paper/2019/file/65fc9fb4897a89789352e211ca2d398f-Paper.pdf}.

\bibitem[Sun et~al.(2020)Sun, Wang, Chen, Ni, Agrawal, Cui, Venkataramani,
  El~Maghraoui, Srinivasan, and Gopalakrishnan]{NEURIPS2020_13b91943}
Xiao Sun, Naigang Wang, Chia-Yu Chen, Jiamin Ni, Ankur Agrawal, Xiaodong Cui,
  Swagath Venkataramani, Kaoutar El~Maghraoui, Vijayalakshmi~(Viji) Srinivasan,
  and Kailash Gopalakrishnan.
\newblock Ultra-low precision 4-bit training of deep neural networks.
\newblock In H.~Larochelle, M.~Ranzato, R.~Hadsell, M.F. Balcan, and H.~Lin,
  editors, \emph{Advances in Neural Information Processing Systems}, volume~33,
  pages 1796--1807. Curran Associates, Inc., 2020.
\newblock URL
  \url{https://proceedings.neurips.cc/paper/2020/file/13b919438259814cd5be8cb45877d577-Paper.pdf}.

\bibitem[Guo et~al.(2020)Guo, Liu, Wang, Han, Li, Lu, and Hu]{9054164}
Jinrong Guo, Wantao Liu, Wang Wang, Jizhong Han, Ruixuan Li, Yijun Lu, and
  Songlin Hu.
\newblock Accelerating distributed deep learning by adaptive gradient
  quantization.
\newblock In \emph{ICASSP 2020 - 2020 IEEE International Conference on
  Acoustics, Speech and Signal Processing (ICASSP)}, pages 1603--1607, 2020.
\newblock \doi{10.1109/ICASSP40776.2020.9054164}.

\bibitem[Yao et~al.(2021)Yao, Dong, Zheng, Gholami, Yu, Tan, Wang, Huang, Wang,
  Mahoney, and Keutzer]{pmlr-v139-yao21a}
Zhewei Yao, Zhen Dong, Zhangcheng Zheng, Amir Gholami, Jiali Yu, Eric Tan,
  Leyuan Wang, Qijing Huang, Yida Wang, Michael Mahoney, and Kurt Keutzer.
\newblock Hawq-v3: Dyadic neural network quantization.
\newblock In Marina Meila and Tong Zhang, editors, \emph{Proceedings of the
  38th International Conference on Machine Learning}, volume 139 of
  \emph{Proceedings of Machine Learning Research}, pages 11875--11886. PMLR,
  18--24 Jul 2021.
\newblock URL \url{https://proceedings.mlr.press/v139/yao21a.html}.

\bibitem[Chen et~al.(2021{\natexlab{b}})Chen, Wang, and Cheng]{Chen_2021_ICCV}
Weihan Chen, Peisong Wang, and Jian Cheng.
\newblock Towards mixed-precision quantization of neural networks via
  constrained optimization.
\newblock In \emph{Proceedings of the IEEE/CVF International Conference on
  Computer Vision (ICCV)}, pages 5350--5359, October 2021{\natexlab{b}}.

\bibitem[Lee et~al.(2019)Lee, Kim, Kang, Shin, Kim, and Yoo]{8481682}
Jinmook Lee, Changhyeon Kim, Sanghoon Kang, Dongjoo Shin, Sangyeob Kim, and
  Hoi-Jun Yoo.
\newblock Unpu: An energy-efficient deep neural network accelerator with fully
  variable weight bit precision.
\newblock \emph{IEEE Journal of Solid-State Circuits}, 54\penalty0
  (1):\penalty0 173--185, Jan 2019.
\newblock ISSN 1558-173X.
\newblock \doi{10.1109/JSSC.2018.2865489}.

\bibitem[He et~al.(2016)He, Zhang, Ren, and Sun]{He_2016_CVPR}
Kaiming He, Xiangyu Zhang, Shaoqing Ren, and Jian Sun.
\newblock Deep residual learning for image recognition.
\newblock In \emph{Proceedings of the IEEE Conference on Computer Vision and
  Pattern Recognition (CVPR)}, June 2016.

\bibitem[Krizhevsky et~al.()Krizhevsky, Nair, and Hinton]{cifar}
Alex Krizhevsky, Vinod Nair, and Geoffrey Hinton.
\newblock Cifar-100 (canadian institute for advanced research).
\newblock URL \url{http://www.cs.toronto.edu/~kriz/cifar.html}.

\bibitem[Deng et~al.(2009)Deng, Dong, Socher, Li, Li, and Fei-Fei]{5206848}
Jia Deng, Wei Dong, Richard Socher, Li-Jia Li, Kai Li, and Li~Fei-Fei.
\newblock Imagenet: A large-scale hierarchical image database.
\newblock In \emph{2009 IEEE Conference on Computer Vision and Pattern
  Recognition}, pages 248--255, June 2009.
\newblock \doi{10.1109/CVPR.2009.5206848}.

\bibitem[Banner et~al.(2018)Banner, Hubara, Hoffer, and
  Soudry]{NEURIPS2018_e82c4b19}
Ron Banner, Itay Hubara, Elad Hoffer, and Daniel Soudry.
\newblock Scalable methods for 8-bit training of neural networks.
\newblock In S.~Bengio, H.~Wallach, H.~Larochelle, K.~Grauman, N.~Cesa-Bianchi,
  and R.~Garnett, editors, \emph{Advances in Neural Information Processing
  Systems}, volume~31. Curran Associates, Inc., 2018.
\newblock URL
  \url{https://proceedings.neurips.cc/paper/2018/file/e82c4b19b8151ddc25d4d93baf7b908f-Paper.pdf}.

\end{thebibliography}

\end{document}